%% file: main.tex
\documentclass[lettersize,journal]{IEEEtran}
\usepackage{amsmath,amsfonts}
\usepackage{algorithmic}
\usepackage{algorithm}
\usepackage{array}
\usepackage[caption=false,font=normalsize,labelfont=sf,textfont=sf]{subfig}
\usepackage{textcomp}
\usepackage{stfloats}
\usepackage{url}
\usepackage{verbatim}
\usepackage{graphicx}
\usepackage{cite}
\usepackage{booktabs}
\usepackage{multirow}
\usepackage{textcomp}
\usepackage{marvosym}

\usepackage{bbding}
\usepackage{color}
\usepackage{colortbl}
\begin{document}

\title{PlanAgent: A Multi-modal Large Language Agent for Closed-loop Vehicle Motion Planning}

\author{Yupeng Zheng$^{\ast\dagger}$, Zebin Xing$^{\ast}$, Qichao Zhang\textsuperscript{\Letter}, Bu Jin, Pengfei Li, Yuhang Zheng, \\ Zhongpu Xia, Kun Zhan, Xianpeng Lang, Yaran Chen, Dongbin Zhao ~\IEEEmembership{Fellow,~IEEE}
\thanks{Yupeng Zheng, Zebin Xin, Qichao Zhang, Bu Jin, Yaran Chen, and Dongbin Zhao are with The State Key Laboratory of Multimodal Artificial Intelligence Systems, Institute of Automation, Chinese Academy of Sciences, Beijing 100190, China, and also with the School of Artificial Intelligence, University of Chinese Academy of Sciences, Beijing 100049, China. 
Zebin Xing is also with the Beijing University of Posts and Telecommunications, Beijing 100876, China.
Yuhang Zheng is with the School of Mechanical Engineering and Automation, Beihang University, Beijing 100191, China.
Pengfei Li is with the Department of Computer Science and Technology, Tsinghua University, Beijing 100084, China.
Zhongpu Xia is an independent researcher. Kun Zhan and Xianpeng Lang are with Li Auto, Beijing, China.
}
\thanks{\textbf{$^{\ast}$ Yupeng Zheng and Zebin Xing contribute equally to this work.}}
\thanks{\textbf{\textsuperscript{\Letter}Qichao Zhang is the corresponding author.}}
\thanks{\textbf{$^{\dagger}$ This work was done when Yupeng Zheng conducted an internship at Li Auto.}}
}

\markboth{This work has been submitted to the IEEE for possible publication}%
{Shell \MakeLowercase{\textit{et al.}}: A Sample Article Using IEEEtran.cls for IEEE Journals}


\maketitle

\input{a0_abstract}
\input{a1_introduction}
\input{a2_related_work}
\input{a3_method}
\input{a4_experimental_setup}
\input{a5_experiments} 
\input{a6_limitation}
\input{a7_conclusion}

\input{main.bbl}

\end{document}

%% file: a0_abstract.tex
\begin{abstract}
Vehicle motion planning is an essential component of autonomous driving technology. 
Current rule-based vehicle motion planning methods perform satisfactorily in common scenarios but struggle to generalize to long-tailed situations. 
Meanwhile, learning-based methods have yet to achieve superior performance over rule-based approaches in large-scale closed-loop scenarios. 
To address these issues, we propose PlanAgent, the first mid-to-mid planning system based on a Multi-modal Large Language Model (MLLM).
MLLM is used as a cognitive agent to introduce human-like knowledge, interpretability, and common-sense reasoning into the closed-loop planning. 
Specifically, PlanAgent leverages the power of MLLM through three core modules. First, an Environment Transformation module constructs a Bird's Eye View (BEV) map and a lane-graph-based textual description from the environment as inputs. 
Second, a Reasoning Engine module introduces a hierarchical chain-of-thought from scene understanding to lateral and longitudinal motion instructions, culminating in planner code generation. 
Last, a Reflection module is integrated to simulate and evaluate the generated planner for reducing MLLM's uncertainty.
PlanAgent is endowed with the common-sense reasoning and generalization capability of MLLM, which empowers it to effectively tackle both common and complex long-tailed scenarios. 
Our proposed PlanAgent is evaluated on the large-scale and challenging nuPlan benchmarks. 
A comprehensive set of experiments convincingly demonstrates that PlanAgent outperforms the existing state-of-the-art in the closed-loop motion planning task.
Codes will be soon released.


\end{abstract}


%% file: a1_introduction.tex
\section{Introduction}

\begin{figure}[t]
\setlength{\abovecaptionskip}{0pt}
\setlength{\belowcaptionskip}{0pt}
    \centering
    \includegraphics[width=\linewidth]{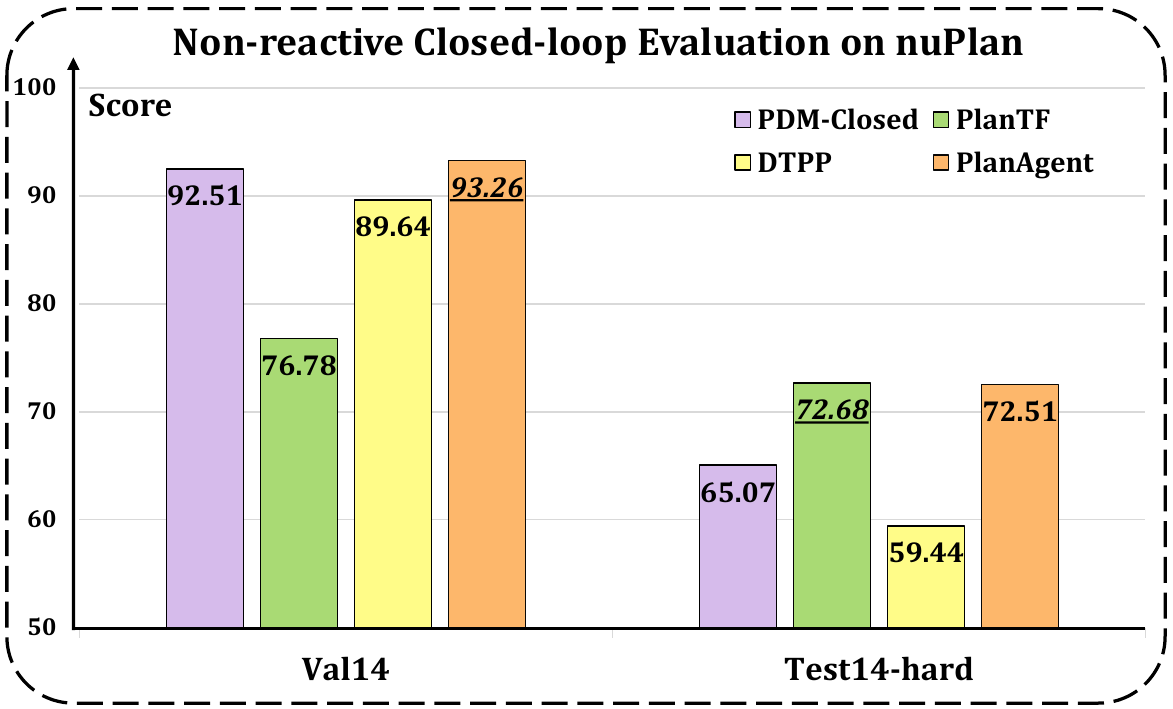}
    \caption{Quantitative results of non-reactive closed-loop motion planning on nuPlan\cite{caesar2021nuplan} Val14 and Test14-hard benchmarks compared with the state-of-the-art rule-based method PDM-Closed\cite{dauner2023parting} and learning-based method PlanTF\cite{cheng2023rethinking} and DTPP\cite{huang2023differentiable}. Our proposed PlanAgent achieves state-of-the-art performance in common scenarios (Val14 benchmark) and demonstrated generalization in more challenging long-tailed scenarios (Test14-hard benchmark). Other methods either perform poorly in common scenarios or find it difficult to generalize to long-tailed scenarios. Please note that \textcolor[RGB]{194, 167, 215}{\textbf{PDM-Closed}}, \textcolor[RGB]{169, 218, 116}{\textbf{PlanTF}}, \textcolor[RGB]{220, 220, 106}{\textbf{DTPP}}, and \textcolor[RGB]{253, 183, 107}{\textbf{PlanAgent}} are denoted by \textcolor[RGB]{194, 167, 215}{\textbf{purple}}, \textcolor[RGB]{169, 218, 116}{\textbf{green}}, \textcolor[RGB]{220, 220, 106}{\textbf{yellow}}, and \textcolor[RGB]{253, 183, 107}{\textbf{orange}}, respectively. The best performances are represented in \underline{\textit{italics and underlined}}.}
    \label{fig: teaser}
\end{figure}

Motion planning for autonomous driving aims to predict the optimal trajectory of the ego vehicle to achieve driving with comfort, safety, and efficiency. 
Recently, it has attracted increasing research interest as one of the key components of autonomous vehicles. 
With advancements in deep learning\cite{9158529, 9284628, li2023conditional, li2023planning}, reinforcement learning\cite{liu2024deep, 9210154, wang2022dynamic, 10530946} and rule-based\cite{thrun2006stanley, bacha2008odin, kesting2007general} techniques, significant progress has been made in autonomous driving planning.
However, due to the complexity and uncertainty of open-world autonomous driving, current rule-based and learning-based methods still face a series of challenges when it comes to high-level closed-loop planning in large-scale scenarios.
As shown in Fig. \ref{fig: teaser}, rule-based methods, such as those noted in \cite{dauner2023parting,treiber2000congested}, have shown adeptness in handling common scenarios but they often struggle with long-tailed situations that demand more complex driving maneuvers. 
Conversely, learning-based approaches frequently grapple with issues of overfitting and long-tailed situations, which leads them to perform worse than the rule-based method\cite{dauner2023parting} in large-scale closed-loop scenarios. 
Moreover, most learning-based methods are grounded in imitation learning in open-loop settings as pointed out in \cite{zhai2023rethinking}, and struggle to generalize to closed-loop scenarios \cite{zhang2022trajgen}.

Recently, the advancements in Large Language Models (LLMs) have opened up new possibilities for autonomous driving planning.
These LLMs, along with their subsequent variants, such as Multi-modal Large Language Models (MLLMs), have demonstrated nearly human-level common-sense reasoning capabilities in a range of domains, including robotics manipulation\cite{huang2023voxposer, liang2023code, li2023manipllm, zheng2024gaussiangrasper, zheng2023enhancing}, multi-modal understanding\cite{wang2023cogvlm, liu2024visual, zhang2023llama} and human-like agent\cite{wang2023voyager, 10423819, 10286969, wei2024editable, chen2024idea}. 
They have become leaders in zero-shot learning and domain adaptation. 
As shown in Tab. \ref{tab: conclusion}, some recent studies\cite{wen2023dilu, sharan2023llm, mao2023language, mao2023gpt} have attempted to integrate the capabilities of LLMs and MLLMs into autonomous driving planning.
However, they encounter the following limitations:
(1) running in simple simulation environments such as Highway-env\cite{wen2023dilu} or open-loop scenarios\cite{mao2023language}, which do not validate the capabilities of LLMs for closed-loop planning in complex real-world scenarios;
(2) using an excessive amount of coordinate numbers as tokens to represent map details\cite{sharan2023llm}, or lacking key information such as lanes\cite{mao2023gpt}, which express scene information inefficiently and harms the performance of LLM for motion planning;
(3) using trajectory waypoints\cite{mao2023gpt} generated by the large language model directly, which introduces safety hazards due to the uncertainty of the LLM.


\begin{table*}
    \centering
    \caption{Comparison between PlanAgent and recent LLM-based approaches.}
    
    \begin{tabular}[b]{c|llllll}
        \toprule
        Attribute  & DiLu\cite{wen2023dilu} & GPT-Driver\cite{mao2023gpt} & Agent-Driver\cite{mao2023language} & LLM-ASSIST\cite{sharan2023llm} & PlanAgent (Ours) \\
        \midrule
         Multi-modal inputs & \XSolidBrush (Text) &  \XSolidBrush (Text) &  \XSolidBrush (Text) &  \XSolidBrush (Text) & \Checkmark (Text \& BEV) \\
         Closed-loop Evaluation & \Checkmark (Closed-loop) & \XSolidBrush (Open-loop) & \XSolidBrush(Open-loop) &\Checkmark (Closed-loop) & \Checkmark (Closed-loop) \\
         Real-world scenarios & \XSolidBrush(Highway-env) & \Checkmark (nuScenes) & \Checkmark (nuScenes) & \Checkmark (nuPlan) & \Checkmark (nuPlan) \\
         GPT Model & \Checkmark (GPT-3.5 \&  GPT-4) & \Checkmark (GPT-3.5) & \Checkmark (GPT-3.5) & \Checkmark (GPT-3.5) & \Checkmark (GPT-4V) \\
         CoT Output &\Checkmark (Action) &\Checkmark(Waypoints) &\Checkmark(Waypoints) & \XSolidBrush(Waypoints or Parameters) & \Checkmark (Hierarchical CoT) \\
        \bottomrule
    \end{tabular}
    \label{tab: conclusion}
\end{table*}

To overcome these challenges and enhance the planner's ability to generalize to diverse real-world scenarios, we propose PlanAgent, the first mid-to-mid planning agent driven by an MLLM.
In particular, PlanAgent leverages the power of MLLM through the following modules:
\textbf{(1) An Environment Transformation} module extracts multi-modal key information from the environment and efficiently converts them to serve as prompt inputs. 
This is achieved by constructing a Bird's Eye View (BEV) map for global positional information and utilizing a lane-graph-based representation to generate textual descriptions for local motion information.
\textbf{(2) A Reasoning Engine} module introduces a hierarchical chain-of-thought to perform multi-round iterated reasoning to instruct the MLLM. 
Through in-context learning, the chain-of-thought bridges high-level scene understanding, intermediate-level longitudinal and lateral motion instructions, and low-level generation of planner code.
\textbf{(3) A Reflection} module that verifies the planner generated by the MLLM through simulation and scoring. 
It aims to filter out unreasonable planners, thereby reducing the impact of LLM's uncertainty on the safety of planning.

The proposed PlanAgent framework effectively enhances the scene understanding capability and planning performance of the MLLM. 
Leveraging our designed approach and the common-sense reasoning ability of LLMs, PlanAgent achieves state-of-the-art performance on both closed-loop reactive and non-reactive settings of nuPlan Val14 and Test14-hard benchmarks.
Detailed ablation studies provide insight into the mechanics and effectiveness of each module.

For convenient reference, our contributions can be summarized as follows:
\begin{itemize}
\item[$\bullet$] We introduce the PlanAgent pipeline, which, to the best of our knowledge, is the first closed-loop mid-to-mid autonomous driving planning agent system based on a Multi-modal Large Language Model (MLLM).

\item[$\bullet$] For efficient scene information representation, we propose an efficient Environment Transformation module that extracts multi-modal information inputs with lane-graph representation.

\item[$\bullet$] For common-sense reasoning and safety planning, we design a Reasoning Engine module that introduces a hierarchical chain-of-thought (CoT) to instruct MLLM to generate planner code and a Reflection module that combines simulation and scoring to filter out unreasonable proposals generated by the MLLM.

\item[$\bullet$] Our approach delivers competitive and generalizable performance on both common Val14 and long-tailed Test14-hard benchmarks of the nuPlan dataset. 
Furthermore, compared to the LLM-based state-of-the-art (SOTA) methods, PlanAgent only requires one-third of the number of tokens needed for a textual description.
\end{itemize}

%% file: a2_related_work.tex
\section{Related Work}
\label{related_work}

\subsection{Motion Planning for Autonomous Driving} 
\textbf{End-to-end} approaches, which utilize raw sensor data as inputs to predict the future trajectory of the ego vehicle directly, have made significant progress in terms of designing network architecture\cite{hu2023planning}, improving intermediate representation\cite{jiang2023vad}, fusing multi-modal information\cite{chitta2022transfuser, ye2023fusionad, chitta2021neat} and incorporating generative models\cite{zheng2024genad} through the utilization of CARLA simulation benchmarks\cite{dosovitskiy2017carla} and nuScenes\cite{caesar2020nuscenes} datasets. 
However, the absence of realistic closed-loop simulators often leads to evaluations being conducted in open-loop settings or confined to simple traffic scenarios within simulators, resulting in a lack of authenticity and diversity necessary for comprehensive evaluation.
  
\textbf{Mid-to-Mid} approaches take recorded real-world perception results and HD Map as input to produce the future trajectory of the ego-vehicle. 
Mid-to-mid planning has significantly progressed thanks to the new release of the large-scale nuPlan dataset and simulation benchmarks. 
The rule-based method PDM\cite{dauner2023parting} achieved SOTA performance on the nuPlan Val14 benchmark by executing intelligent driving models (IDM\cite{treiber2000congested}) with varying parameters and evaluating them through simulation. 
However, PDM struggles to generalize to long-tail scenarios, underperforming on the subsequent long-tail benchmark Test14-hard\cite{cheng2023rethinking}. 
Learning-based methods\cite{scheel2022urban, hallgarten2023prediction, huang2023gameformer, hu2023imitation, huang2023differentiable} like GC-PGP\cite{hallgarten2023prediction} transforms high-performance prediction models into goal-oriented planners using the representation of lane graphs. 
Gameformer\cite{huang2023gameformer} utilizes the DETR\cite{carion2020end} architecture, and Hoplan\cite{hu2023imitation} predicts future trajectories by forecasting spatial-temporal occupancy heatmaps. 
These methods typically incorporate post-optimizers to enhance the reliability of produced trajectories. 
DTPP\cite{huang2023differentiable} introduces a differentiable framework for the joint training of ego-conditioned prediction and cost models.
PlanTF\cite{cheng2023rethinking} introduces an attention-based state dropout encoder (SDE) and augmentation techniques, effectively mitigating compound errors. 
Moreover, based on the nuPlan test split, PlanTF proposes a long-tail scenario benchmark Test14-hard for validating model generalization in long-tailed scenarios. 
Our research is grounded in the mid-to-mid closed-loop setting, delving into the common sense reasoning and generalization capabilities of multi-modal large language models in vehicle motion planning.

\subsection{Large Language Models for Autonomous Planning} 
The LLM has demonstrated exceptional common-sense reasoning and zero-shot generalization capabilities. 
Its subsequent variant, the MLLM, further incorporates encoding for additional modalities, enabling the understanding of multi-modal data. 
On the one hand, some research endeavors have attempted to integrate LLMs and MLLMs in vision-centric autonomous driving to enhance interpretability\cite{jin2023adapt,jin2024tod3cap} and generalization capabilities\cite{Li2024DrivingEW, 10492662, Pan2024VLPVL}.
Agent-Driver\cite{mao2023language} employs an LLM as an agent to invoke perception tools to acquire scene information from raw images for reasoning and predicting trajectories.
These methods use temporal image sequences or LiDAR as input to produce vehicle motion commands, control signals, or predict future trajectories in open-loop environments, ignoring issues such as cumulative errors and tracking errors. 
Based on the CARLA\cite{dosovitskiy2017carla} simulator, LMDrive\cite{shao2023lmdrive} has implemented the first LLM-based closed-loop end-to-end autonomous driving system.
DriveLLM\cite{10297415} presents a novel framework that integrates LLMs with existing autonomous driving modules, enabling informed decision-making in real-world scenarios.

\begin{figure*}[t]
\setlength{\abovecaptionskip}{0pt}
\setlength{\belowcaptionskip}{0pt}
    \centering
    \includegraphics[width=\linewidth]{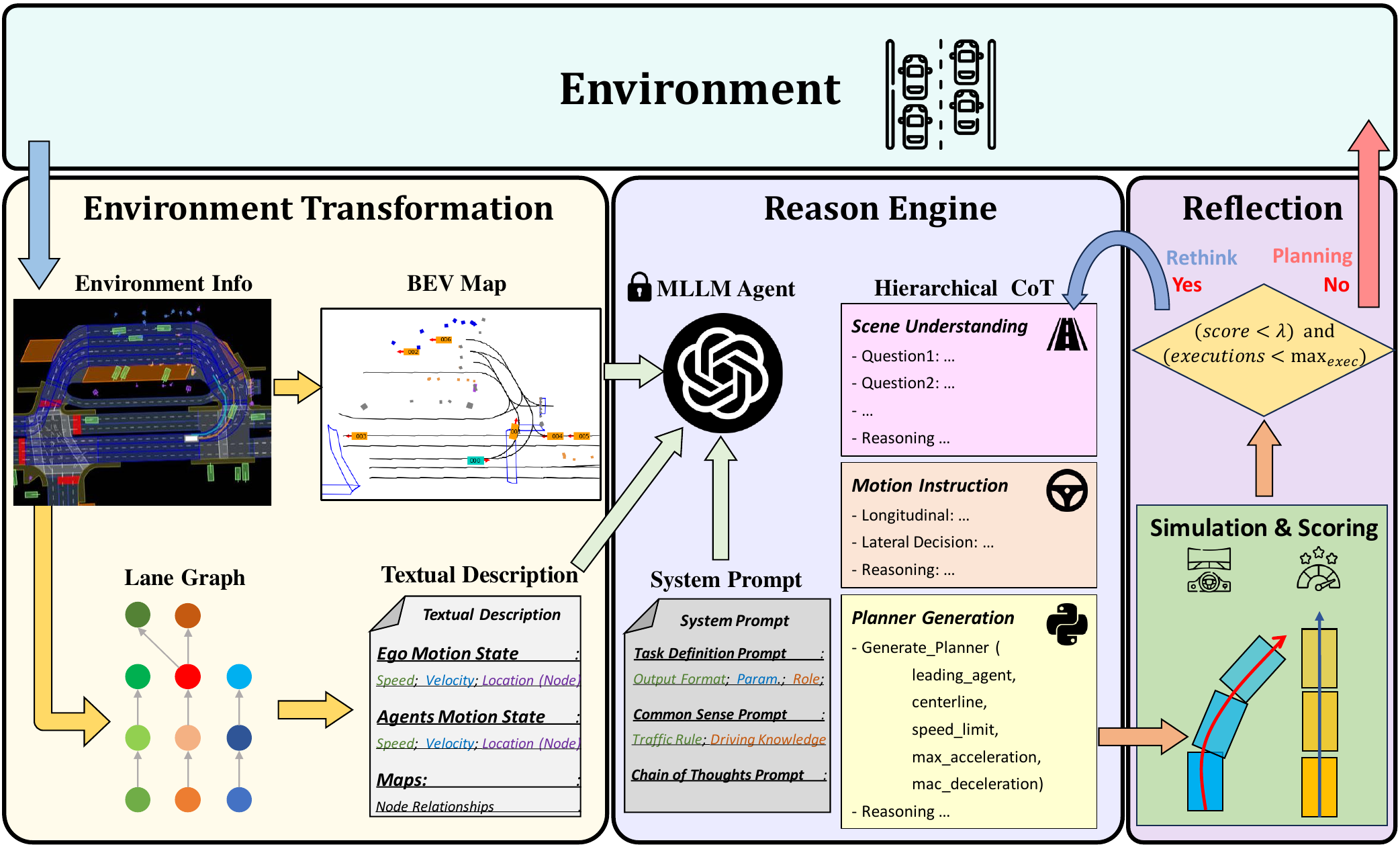}
    \caption{Based on a MLLM, we propose a novel planning agent pipeline comprising three modules: Environment Transformation, Reasoning Engine, and Reflection Module. In the Environment Transformation module, key information about the environment is extracted to form a BEV map and construct a lane-graph representation. Subsequently, the lane graph is translated into textual descriptions and used as scenario prompts along with the BEV map. In the Reasoning Engine module, an MLLM generates planner codes based on the IDM\cite{treiber2000congested} planner through hierarchical chain-of-thought reasoning with scenario prompts and pre-defined system prompts (including task definition prompts, common sense prompts, and chain-of-thought guidance prompts). In the Reflection module, the planner generated by Reason Engine is simulated and evaluated. Whether to execute or rethink depends on the assessed score.}
    \label{fig: pipeline}
\end{figure*}

On the other hand, some recent research \cite{wen2023dilu,sharan2023llm,mao2023gpt,sha2023languagempc} have applied LLM in mid-to-mid autonomous driving planning. 
DiLu\cite{wen2023dilu} introduced a closed-loop framework paradigm for LLMs that includes reasoning, reflection, and memory modules that incorporate human knowledge. 
GPT-driver\cite{mao2023gpt} achieves motion planning through natural language modeling by converting scenes into language and fine-tuning GPT. 
LLM-ASSIST\cite{sharan2023llm} employs LLMs to generate controller parameters.
Unlike these methodologies, we propose a planning agent framework based on MLLM. 
Given BEV map and lane-graph-based scene descriptions extracted from the environment, we design a hierarchical chain-of-thought from perception to longitudinal and lateral motion instruction and planner code generation. 
Utilizing this in-context learning process, we effectively integrate multi-modal large language models into autonomous driving planning tasks, leveraging their capabilities to enhance planning reliability and safety.
Moreover, unlike existing LLM-based approaches that directly convert scenes into textual inputs, we propose a more efficient Environment Transformation module that reduces the usage of the tokens to describe the scenario. 

%% file: a3_method.tex
\section{Methodology}
\label{method}

\subsection{Overall Architecture}
Inspired by Large Language Model agents, we have introduced a novel and effective autonomous driving framework named PlanAgent, which utilizes an MLLM as a vehicle motion planning agent. 
As depicted in Fig. \ref{fig: pipeline}, the architecture of PlanAgent includes three core modules: Environment Transformation module, Reasoning Engine module, and Reflection module.
At a given planning interval, PlanAgent retrieves scene information through the Environment Transformation module, calls the Reasoning Engine to generate a planner and finally verifies the planner using the Reflection module.

Specifically, we first introduce how the Environment Transformation module efficiently extracts key information from the environment to construct a BEV map and lane-graph-based textual description as scenario prompts in section \ref{subsec: ET}.
Next, in section \ref{subsec: RE}, we elaborate on how to define the system prompt and how to design a hierarchical chain-of-thought in Reasoning Engine to instruct the large model through in-context learning to understand the scene, give instructions, and generate code of a planner.
Finally, in section \ref{subsec: RM}, we introduce how the Reflection module examines the generated planners from the MLLM to mitigate the impact of the MLLM's uncertainty on planning safety.
The following sections will elaborate on implementing Environment Transformation, Reasoning Engine, and Reflection modules.

\subsection{Environment Transformation}
\label{subsec: ET}
The quality of prompts is a key factor in the quality of output produced by LLMs.
To efficiently obtain prompts from complex environment data to be consistent with the semantic space of LLM, we design an Environment Transformation module.
It divides the scene information into two parts: global and local. 
The global information represents the scenario type, providing semantic priors for vehicle motion planning, such as the necessity to exercise caution with vehicle merging at roundabouts, the prohibition against lane-changing on single-lane roads, and so on. 
The local information represents the motion of the ego-vehicle and surrounding agents, directly affecting the vehicle's longitudinal and lateral action planning. 
Specifically, the Environment Transformation module first extracts essential scene context information and converts it into a BEV map as global semantic information.
Next, it constructs a lane-graph to record maps and related agents.
The lane-graph will be converted into textual description as local motion information.
To simultaneously introduce global semantic information and local motion information, we use BEV map and textual description to serve as multi-modal scenario prompts.

\textbf{BEV Map:} We first extract map information, agent information, and obstacle information from the environment. 
Map information includes (1) centerlines, (2) crosswalks, and (3) lanes. 
Agent information covers the (4) ego-vehicle, (5) other vehicles, (6) bicycles, and (7) pedestrians.
Obstacle information represents (8) static obstacles. 
The eight road elements mentioned above are visualized with different colors when constructing the BEV map to enhance the multi-modal large model's comprehension of the road structure.
To illustrate the scene's dynamic and stationary information, each agent is marked with a red arrow pointing in the direction of movement.
The length of the arrow symbolizes the agent's speed as shown in the left side of Fig. \ref{fig: pipeline}.

\begin{figure}
\setlength{\abovecaptionskip}{0pt}
\setlength{\belowcaptionskip}{0pt}
    \centering
    \includegraphics[width=\linewidth]{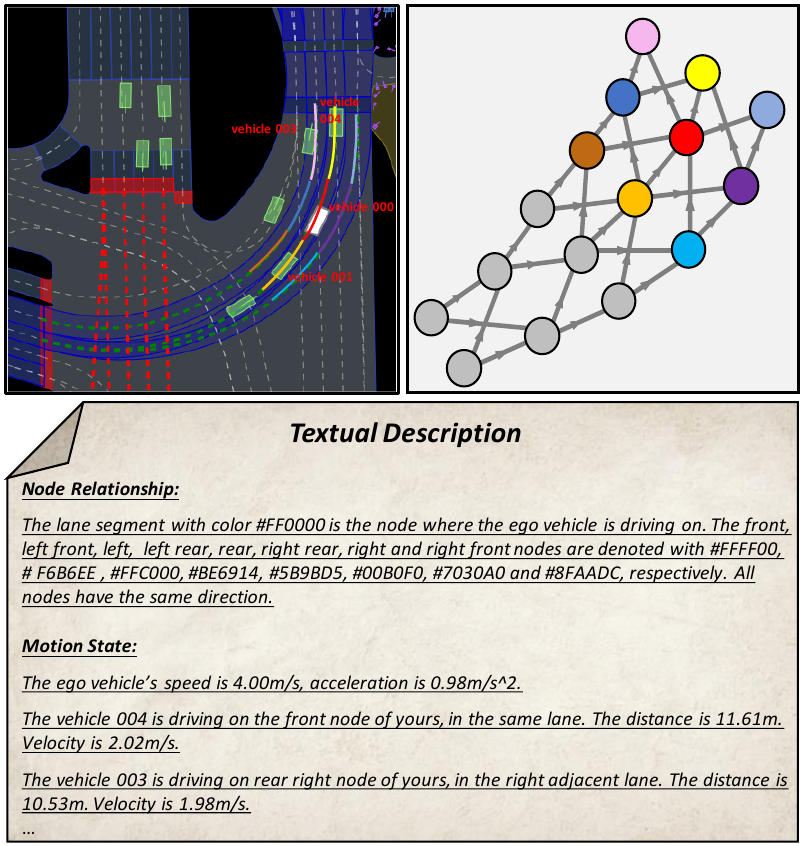}
    \caption{The top of the picture shows the process of constructing a lane map (top right) based on the environment (top left). The white square on the left represents the ego vehicle. The red node on the right indicates the centerline segment where the ego vehicle is located, while nodes of other colors correspond to lane segments of the same color on the left. The bottom of the picture displays the converted text description of the scenario based on the lane-graph, including node relationships and motion states.}
    \label{fig: lane}
\end{figure}

\textbf{Textual Description:} Previous methods\cite{mao2023gpt, sharan2023llm} that deliver all agent information and lane information into LLMs in the form of numerical coordinates are redundant and inefficient.
Furthermore, an abundance of numerical coordinates might not properly align with the semantic space of LLM, subsequently diminishing its performance and its ability to understand scenes effectively.
To effectively align map lane information with the semantic space of LLM, we represent map lanes as a set of lane nodes in the form of a lane-graph.
The positional and connection relationships between nodes provide a good semantic and geometry representation of the relationships between lanes. 
Specifically, in order to ensure that the length of each node is the same in complex road situations, we divide the centerlines of lanes into segments of equal length, using these segments as nodes in the lane-graph.
As shown in Fig. \ref{fig: lane}, for a three-lane scene, each node naturally forms eight types of adjacent nodes in different directions. 
We convert the positional and connection information between nodes into textual description, efficiently representing complex real-world map lane information. 
To efficiently convey information about other agents and obstacles, we retain only the agents and obstacles on the eight-lane nodes around the node where the ego-vehicle is located. 
We obtain their speed, distance from the ego-vehicle, and the located lane node from the environment and then translate this information into textual description as shown in the bottom of Fig. \ref{fig: lane}.
It is notable that we also describe the status of traffic lights. 

The Environment Transformation module develops both a BEV map for global information and a lane-graph-based textual description for local information from the environment. These serve as scenario prompts for the MLLM employed by the subsequent Reasoning Engine.

\subsection{Reasoning Engine}
\label{subsec: RE}
How to introduce the power of MLLMs into the autonomous driving planning process to achieve a planning system capable of common-sense reasoning and generalization remains an open area of exploration.
In this section, we propose the Reasoning Engine for autonomous driving planning, as shown on the middle side of Fig. \ref{fig: pipeline}. 
It takes the scenario prompts constructed by the Environment Transformation module and predefined system prompts as inputs. 
Next, an MLLM generates the planner code based on the IDM\cite{treiber2000congested} planner through multiple rounds of reasoning under the guidance of the hierarchical chain-of-thought.
In this process, the reasoning ability of the MLLM is introduced into the driving planning task through in-context learning. 
The formulation of the predefined system prompt and the design of the hierarchical chain-of-thought are expanded below:

\begin{figure}
\setlength{\abovecaptionskip}{0pt}
\setlength{\belowcaptionskip}{0pt}
    \centering
    \includegraphics[width=\linewidth]{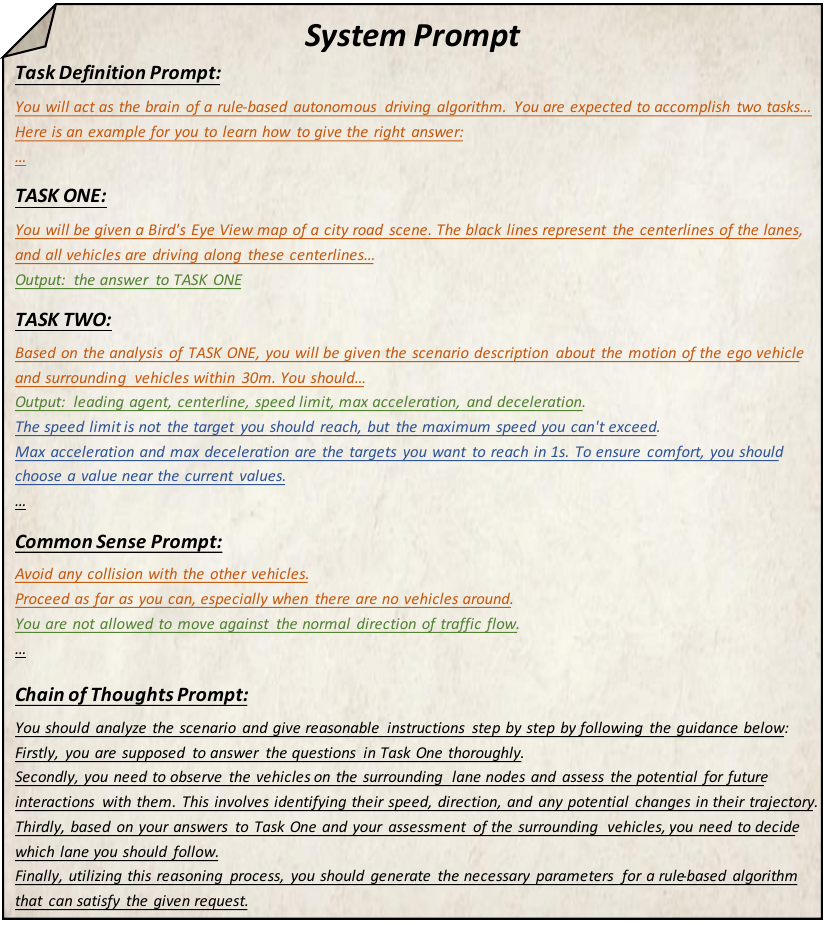}
    \caption{The detailed example of the system prompt for PlanAgent. It consists of a task definition prompt, a common sense prompt, and a chain-of-thought prompt. }
    \label{fig: lane}
\end{figure}

\textbf{System Prompt Formulation:} The prompts provided to the reasoning engine are divided into two parts: scenario prompts used to describe the scene and system prompts used to define tasks, introduce driving common sense, and instruct hierarchical reasoning. 
Since the MLLM that has not been fine-tuned with autonomous driving data can not execute accurate reasoning when handling complex closed-loop planning tasks, the system prompt needs to feed the task definition, format, and a reasonable case of planner generation information to the MLLM in the form of in-context learning. 
As shown in the middle of Fig. \ref{fig: pipeline}, the predefined system prompts consist of a task definition prompt, a common sense prompt, and a chain-of-thought prompt.

Specifically, the task definition prompt represents the role of the MLLM in the planning task. 
Its goal is to standardize the format of input data, give demonstration cases of planner code generation and clarify the meaning of the generated code, that is, the physical meaning of each parameter in the IDM planner of the generated controller, including the followed centerline ($c$), the leading agent ($la$), driving speed limit ($v_0$), maximum acceleration ($acc$) and maximum deceleration ($dec$). 
The common sense prompt is made up of the basic knowledge and traffic rules usually needed for safe driving on the road. 
The chain-of-thought prompt aims to instruct the MLLM to understand the scene, give motion instructions, and generate planner codes step by step. 
To achieve this goal, we design a series of questions to provide more guidance.
Details of the hierarchical chain-of-thought are explained in the next section.

\textbf{Hierarchical Chain-of-thought:} As shown in the middle of Fig. \ref{fig: pipeline}, hierarchical chain-of-thought includes scene understanding, motion instruction, and planner generation.

{Scene Understanding: }
By analyzing PDM-Closed\cite{dauner2023parting}, we find that different types of scenarios require different planners. 
In particular, IDM planners with different parameters have large performance differences in different types of scenarios.
Therefore, understanding the global and local information of the scenarios is crucial for the MLLM to generate the right planner.
To this end, we design global understanding questions such as the type of the scenario, traffic light status, precautions in this type of scene, and local understanding questions such as the status of the ego-vehicle, whether there are other vehicles in the same lane, etc.
These questions provide guidance to the MLLM to understand the scenario.

{Motion Instruction Production:}
Next, according to the analysis of this scenario, we ask the MLLM to provide the longitudinal and lateral motion instruction for the ego-vehicle and give the rationale behind the instruction.
The longitudinal instruction includes accelerating, decelerating, and maintaining the current speed, and the lateral instruction includes changing to the left lane, changing to the right lane, and keeping the current lane.

{Planner Code Generation: }
After analyzing the scenario and providing the motion instruction for the ego-vehicle, the MLLM learns how to generate planner python code by understanding the physical meaning of each parameter in the IDM planner and learning the demonstration planner case in the system prompt.
When given a centerline $c$ and the leading agent $la$, the distance to the leading agent $s$, and the current velocity $v$, IDM generates the longitudinal acceleration as following:
\begin{equation}
    a = \textcolor[RGB]{255, 0, 0}{acc}\Bigg(1-\bigg(\frac{v}{\textcolor[RGB]{255, 0, 0}{v_0}}\bigg)^\delta - \bigg(\frac{s^*}{s}\bigg)^2 \Bigg),
    \label{eq:idm}
\end{equation}
\begin{equation}
    \frac{dv}{dt} = min(acc,max(a,\textcolor[RGB]{255, 0, 0}{dec}))
\end{equation}
where $v_0$ is the speed limit, $acc$ is the acceleration limit and $dec$ is the deceleration limit. 
$v_0$, $acc$, $dec$ and $c$ are four important hyperparameters that directly affect longitudinal and lateral motion.
Based on the IDM planner, the MLLM generates the Python code which calls IDM planner with different hyperparameters as follows:
\begin{equation}
Generate\_IDM\_Planner(c, la, v_0, acc, dec).
\end{equation}

\subsection{Reflection}
\label{subsec: RM}
For autonomous driving planning tasks, safety is crucial. 
Inspired by the human decision-making principle of "\textit{Look before you leap}," we design a Reflection module in our PlanAgent system, as depicted on the right side of Fig. \ref{fig: pipeline}. 
This module intends to mitigate the effects of uncertainty from the MLLM in the Reasoning Engine, thereby enhancing the safety of closed-loop planning.

In particular, we integrate the simulation process proposed by the PDM-Closed\cite{dauner2023parting} method with the planner generation of the MLLM agent. 
Every planner produced by the Reasoning Engine undergoes simulation, and a simulated driving score (denoted as $s$) is derived based on metrics such as potential for collision and comfort levels. 
If the simulated driving score drops below a specific threshold (denoted as $\lambda$), indicating an incorrect decision and planner generation, the Reasoning Engine will be notified to reprocess the chain-of-thought and create a new planner. The process is formulated as
\begin{equation}
\rm{Decision} = \left\{  
         \begin{array}{lr}  
         {\rm{Planning}} , & s \geq \lambda \\  
         {\rm{Rethinking}} , & s \ \textless \ \lambda \\
         \end{array}  
\right.
\end{equation}
Unlike the long-term simulation deployed in PDM, we employ short-term simulation, intending to minimize the impact of uncertainty with each invocation of the MLLM. 
Moreover, since PDM simulation utilizes a uniform speed model to predict the future motion of other agents, our short-term simulation can minimize the cumulative error in predicting dynamic agents' motion. 
This results in a more accurate simulation score.
Lastly, it is notable that due to decision-making time constraints, the Reflection module is set to perform a maximum number of executions (designated as $\rm{max}_{exec}$). 
Once this limit is exceeded, the last generated controller code is selected as the final output.

%% file: a4_experimental_setup.tex
\section{Experimental Setup}
\label{setup}

\subsection{Implementation Details}
\textbf{MLLM Agent:}
PlanAgent utilizes a pre-trained Multi-moda Large Language Model and can be applied to planning tasks without training.
Specifically, in this paper, we choose GPT-4V as the MLLM agent.
Other MLLM experiments will be shown in the Sec. \ref{exp_MLLM}.

\textbf{Hyperparameters:}
The threshold $\lambda$ is set to $0.75$ and the maximum number of executions $\rm{max}_{exec}$ is set to 3.

\subsection{Datasets}
nuPlan\cite{caesar2021nuplan} is a large-scale closed-loop planning benchmark for autonomous driving, which consists of 1500h human driving data from 4 cities: Boston, Pittsburgh, Las Vegas, and Singapore. These cities are across the US, posing challenges for planners to identify and adapt to the varying traffic patterns. In our experiments, we exploit the nuPlan Val14 and nuPlan Test14-hard to investigate the effectiveness of our PlanAgent.
\textbf{nuPlan Val14} is the original validation set of nuPlan.
\textbf{nuPlan Test14-hard} represents the long-tailed scenarios in nuPlan, defined in PlanTF\cite{cheng2023rethinking}.

\subsection{Metrics}
We use the official closed-loop evaluation metrics non-reactive closed-loop score (NR-CLS) and reactive closed-loop score (R-CLS) officially provided by nuPlan\cite{caesar2021nuplan} as our evaluation metrics.
R-CLS and NR-CLS have the same calculation method. 
The ``reactive'' means the background traffic is controlled by the IDM\cite{treiber2000congested}, while the ``non-reactive'' utilizes the log-reply of other agents.
They measure the actual driving performance in simulation, including (1) traffic rule violations, like collision, off-road trajectories, and speed limit; (2) human driving similarities, like longitudinal velocity error, longitudinal stop position error, or lateral position error; (3) vehicle dynamics, like comfort or feasibility; (4) goal achievement, progress along the experts' route. 
Both NR-CLS and R-CLS range from 0 to 100, where higher scores are better.

\subsection{Baseline Methods}
We compare state-of-the-art methods with our PlanAgent, including \textbf{(1) RasterModel}\cite{caesar2021nuplan}, which utilizes a CNN to encode the raster and generate the future trajectory of the ego car; \textbf{(2) UrbanDriver}\cite{scheel2022urban}, which introduces a PointNet-like architecture to encode the input elements (state observation) and employs transformer layers for global reasoning; \textbf{(3) GameFormer}\cite{huang2023gameformer}, which is based on transformer encoder-decoder architecture and formulates the planning as level-k game; \textbf{(4) PlanTF}\cite{cheng2023rethinking}, which introduces state dropout encoder (SDE) to address the shortcut learning problems in planning; \textbf{(5) DTPP}\cite{huang2023differentiable}, which employs a differentiable joint training method for  both ego-conditioned prediction and cost models; \textbf{(6) PDM}\cite{dauner2023parting}, a well designed rule-based architecture that extend IDM with different hyper-parameters; and \textbf{(7) LLM-ASSIST}\cite{sharan2023llm}, employing a LLM to choose the better hyper-parameters for PDM.

%% file: a5_experiments.tex
\section{Experiments}

\label{exp}
\subsection{Comparison with SOTA Methods}
\begin{table*}[]
\centering
\setlength{\tabcolsep}{10pt}
\renewcommand{\arraystretch}{1.2}
\caption{Comparison with Competitive Methods on Val14 and Test14-hard benchmarks of nuPlan Closed-Loop Planning Challenge.}
\begin{tabular}[b]{ccc|cc|cc}

\toprule
\multicolumn{3}{c}{Planners} & \multicolumn{2}{c}{Val14} & \multicolumn{2}{c}{Test14-hard} \\ \midrule
Category & \multicolumn{1}{c}{Method} & \multicolumn{1}{c|}{Publication} & NR-CLS $\uparrow$ & \multicolumn{1}{c|}{R-CLS $ \uparrow$ } & NR-CLS $ \uparrow$ & \multicolumn{1}{c}{R-CLS $\uparrow$}  \\ \midrule
Expert & \multicolumn{1}{c}{Log-replay} & - & 94.03 & \multicolumn{1}{c|}{75.86} & 85.96 & \multicolumn{1}{c}{68.80} \\ \midrule
\multirow{2}{*}{Rule-based} & \multicolumn{1}{c}{IDM~\cite{treiber2000congested}} & Physical Review E & 70.39 & \multicolumn{1}{c|}{72.42} & 56.16 & 62.26  \\
 & \multicolumn{1}{c}{PDM-Closed~\cite{dauner2023parting}} & \multicolumn{1}{c|}{CoRL 2023} & 92.51 & \multicolumn{1}{c|}{{91.79}} & 65.07 & \underline{75.18} \\ \midrule
\multirow{7}{*}{Learning-based} & \multicolumn{1}{c}{RasterModel~\cite{caesar2021nuplan}} & \multicolumn{1}{c|}{CVPR 2021} & 69.66 & \multicolumn{1}{c|}{67.54}  & 49.47 & \multicolumn{1}{c}{52.16} \\
 & \multicolumn{1}{c}{UrbanDriver~\cite{scheel2022urban}} & \multicolumn{1}{c|}{CoRL 2022} & 63.27 & \multicolumn{1}{c|}{61.02} & 51.54 & 49.07 \\
 & \multicolumn{1}{c}{GC-PGP~\cite{hallgarten2023prediction}} & \multicolumn{1}{c|}{ITSC 2023} & 55.99 & \multicolumn{1}{c|}{51.39} & 43.22 & 39.63 \\
 & \multicolumn{1}{c}{PDM-Open~\cite{dauner2023parting}} & \multicolumn{1}{c|}{CoRL 2023} & 52.80 & \multicolumn{1}{c|}{57.23} & 33.51 & 35.83 \\
 & \multicolumn{1}{c}{GameFormer~\cite{huang2023gameformer}} & \multicolumn{1}{c|}{ICCV 2023} & 80.80 & \multicolumn{1}{c|}{79.31}  & 66.59 & 68.83 \\
 & \multicolumn{1}{c}{PlanTF~\cite{cheng2023rethinking}} & \multicolumn{1}{c|}{ICRA 2024} &  84.83 & \multicolumn{1}{c|}{76.78}  & \textbf{72.68} & 61.70 \\ 
  & \multicolumn{1}{c}{DTPP~\cite{huang2023differentiable}} & \multicolumn{1}{c|}{ICRA 2024} &  89.64 & \multicolumn{1}{c|}{89.78}  & 59.44 & 62.94 \\ \midrule
 \multirow{3}{*}{LLM}  & \multicolumn{1}{c}{LLM-ASSIST$\rm{_{UNC}}$*~\cite{sharan2023llm}} & arXiv 2024 & {90.11} & \multicolumn{1}{c|}{90.32}  & - & - \\  & \multicolumn{1}{c}{LLM-ASSIST$\rm{_{PAR}}$*~\cite{sharan2023llm}} & arXiv 2024 & \underline{93.05} & \multicolumn{1}{c|}{\underline{92.20}}  & - & - \\
 & \multicolumn{1}{c}{PlanAgent(Ours)} & - & {\textbf{93.26}} & \multicolumn{1}{c|}{\textbf{92.75}}  & \underline{72.51} & \textbf{76.82} \\ 
 \bottomrule 
 \multicolumn{7}{l}{* Since the LLM-ASSIST\cite{sharan2023llm} is not open-source, we only report its results publicly available on Val14 benchmark.}\\
 \multicolumn{7}{l}{The best and second-best performances are represented by \textbf{bold} and \underline{underline} respectively.}

\end{tabular}
\label{tab: main_result}
\end{table*}

\begin{figure}[t]
\setlength{\abovecaptionskip}{0pt}
\setlength{\belowcaptionskip}{0pt}
    \centering
    \includegraphics[width=\linewidth]{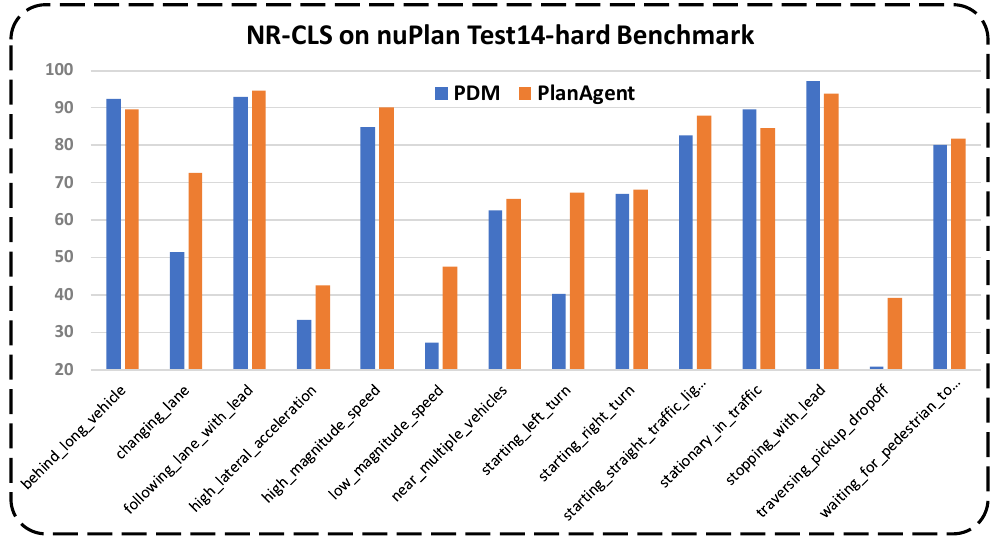}
    \caption{The comparison of the NR-CLS metric between our proposed PlanAgent and PDM-Closed\cite{dauner2023parting} across 14 scenario types based on the nuPlan Test14-hard benchmark.}
    \label{fig: statisc}
\end{figure}

In this section, we first compare the proposed PlanAgent with three categories of state-of-the-art methods: rule-based, learning-based, and LLM-based approaches in both closed-loop reactive and closed-loop non-reactive settings on the nuPlan Val14 and Test14-hard benchmarks. The specific quantitative results are presented in Tab \ref{tab: main_result}. 
Our proposed PlanAgent delivers competitive and generalizable performance compared to other methods.

\textbf{Competitive Results:} In the common scenarios of Val14 benchmark, PlanAgent outperforms other rule-based, learning-based, and large language model-based methods regarding NR-CLS and R-CLS metrics. 
In the more challenging long-tailed scenarios of the Test14-hard benchmark, PlanAgent achieves a higher R-CLS score than the other three categories of state-of-the-art methods, with only a slightly lower NR-CLS score compared to planTF\cite{cheng2023rethinking}. 
The outstanding performance in closed-loop scenarios validates the effectiveness of PlanAgent in closed-loop autonomous driving planning tasks, thereby enhancing the safety of autonomous driving planning systems.
In addition, Fig. \ref{fig: statisc} shows the comparison of the NR-CLS metric between our proposed PlanAgent and PDM-Closed\cite{dauner2023parting} across 14 scenario types based on the nuPlan Test14-hard benchmark. 
PlanAgent performs better in \textit{changing\_lane}, \textit{turn}, \textit{high\_lateral\_acceleration}, \textit{traversing\_pickup\_dropoff} and other scenarios that require complex driving maneuvers, demonstrating the effectiveness of PlanAgent in complex scenarios.

\textbf{Generalizable Results:} The rule-based method PDM-Closed\cite{dauner2023parting} performs well on Val14 benchmark but demonstrates limited generalization on the long-tailed scenarios of Test14-hard benchmark. 
The learning-based method planTF\cite{cheng2023rethinking} achieves the highest NR-CLS score in long-tailed scenarios but its performance significantly drops in regular scenarios. 
Compared to the aforementioned methods, our proposed PlanAgent demonstrates good performance on both Val14 and Test14-hard, indicating that the generalization and reasoning capabilities of multi-modal large language models are successfully transferred to the autonomous driving planning system, thereby enhancing its generalization across diverse driving scenarios.

\textbf{Tokens Utilization:} We then compare the average number of tokens used for the scene description of PlanAgent with other LLM-based methods, namely GPT-driver\cite{mao2023gpt} and LLM-ASSIST\cite{sharan2023llm} in Tab. \ref{tab: token}. 
A lower average token usage represents a more efficient scene description. 
To ensure a fair comparison, the scene description prompts of DiLu and GPT-driver are adapted for nuPlan scenarios. 
The token usage of scene description prompts in LLM-ASSIST is counted based on the sample prompts provided in the paper. 
As shown in Tab. \ref{tab: token}, our proposed PlanAgent has the lowest average token usage, achieving SOTA performance while describing scenes more efficiently compared to other LLM-based methods due to the designed lane-graph-based textual description.

\begin{table}
    \centering
    \caption{The average number of tokens used to describe the scene in the textual description.}
    \resizebox{0.9\columnwidth}{!}{
    \begin{tabular}[b]{c|ccc}
        \toprule
        Method  & GPT-Driver{$\dagger$} & LLM-ASSIST* & PlanAgent (Ours) \\
        \midrule
         Tokens $\downarrow$  & 448.66 & 425.81 & \textbf{141.32} \\
        \bottomrule
        \multicolumn{4}{l}{} \\
        \multicolumn{4}{l}{* denotes the result based on the sample prompts provided in the paper.} \\
        \multicolumn{4}{l}{$\dagger$ denotes the results adapted for nuPlan scenarios.}
    \end{tabular}
    }
    \label{tab: token}
\end{table}

\subsection{Ablation Study}
We conducted a detailed analysis of the effectiveness of each component of PlanAgent on the nuPlan Test14-hard benchmark. 
Tab. \ref{tab: ablation_et} presents the effectiveness of each component in the Environment Transformation module, while Tab. \ref{tab: ablation_rr}  shows the effectiveness of each component in the Reason Engine and the Reflection module.

\textbf{Ablation Study of Environment Transformation Module.}
The comparisons in the first and third rows of Tab. \ref{tab: ablation_et} as well as the second and third rows, demonstrate the effectiveness of adding BEV map modality input and utilizing lane-graph representation in the Environment Transformation module. 
They result in score improvements of 1.5 and 1.3 in NR-CLS, and 1.2 and 1.6 in R-CLS, respectively.

\begin{table}[]
    \centering
    \setlength{\tabcolsep}{0.02\linewidth}
    \caption{Ablation study of Environment Transformation Module on Test14-hard benchmark.}
    \begin{tabular}[b]{ccc|cc}
        \toprule
        Row & 
        BEV Map & 
        Lane-graph &
        NR-CLS $\uparrow$ &
        R-CLS $\uparrow$ \\
        \midrule
        1 & $\times$ & $\times$ & 67.96 & 72.44 \\
        2 & $\times$ & \checkmark & 70.97 & 75.61 \\
        3 & \checkmark & $\times$ & 71.23 & 75.25 \\
        4 & \checkmark & \checkmark & 72.51 & 76.82 \\
        \bottomrule
    \end{tabular}
    \label{tab: ablation_et}
\end{table}

\textbf{Ablation Study of Reasoning Engine Module and Reflection Module.}
The comparisons in the first and fourth rows of Tab. \ref{tab: ablation_rr}, as well as the second and fourth rows, demonstrate the efficacy of high-level scene understanding and mid-level motion instruction generation in the hierarchical chain-of--thought in the Reason Engine on the generation of planner code in PlanAgent. 
The absence of in-context learning for scene understanding and mid-level motion instruction generation in the prompt leads to a decrease of 2.42 and 1.91 in NR-CLS scores, respectively.
The comparison in the third and fourth rows of Tab. \ref{tab: ablation_rr} validates the importance of the Reflection module for safe closed-loop planning. 
The absence of the Reflection module, which performs safety checks on the planner generated by the MLLM, leads to a significant decrease in planning performance, with a reduction of 2.67 in NR-CLS score, illustrating the impact of uncertainty in the MLLM on the safety of the planning system.

\begin{table}[]
    \centering
    \setlength{\tabcolsep}{0.01\linewidth}
    \caption{Ablation study of Reason Engine and Reflection Module on Test14-hard benchmark.}
    \begin{tabular}[b]{cccc|cc}
        \toprule
        Row & 
        Scene Understanding & 
        Instruction &
        Reflection &
        NR-CLS $\uparrow$ &
        R-CLS $\uparrow$ \\
        \midrule
        1 & $\times$ & \checkmark & \checkmark & 70.09 & 75.26 \\
        2 & \checkmark & $\times$ & \checkmark & 70.60 & 75.39 \\
        3 & \checkmark & \checkmark & $\times$ & 69.84 &  74.08 \\
        4 & \checkmark & \checkmark & \checkmark & 72.51 & 76.82 \\
        \bottomrule
    \end{tabular}
    \label{tab: ablation_rr}
\end{table}

\begin{figure*}[]
\setlength{\abovecaptionskip}{0pt}
\setlength{\belowcaptionskip}{0pt}
    \centering
    \includegraphics[width=\linewidth]{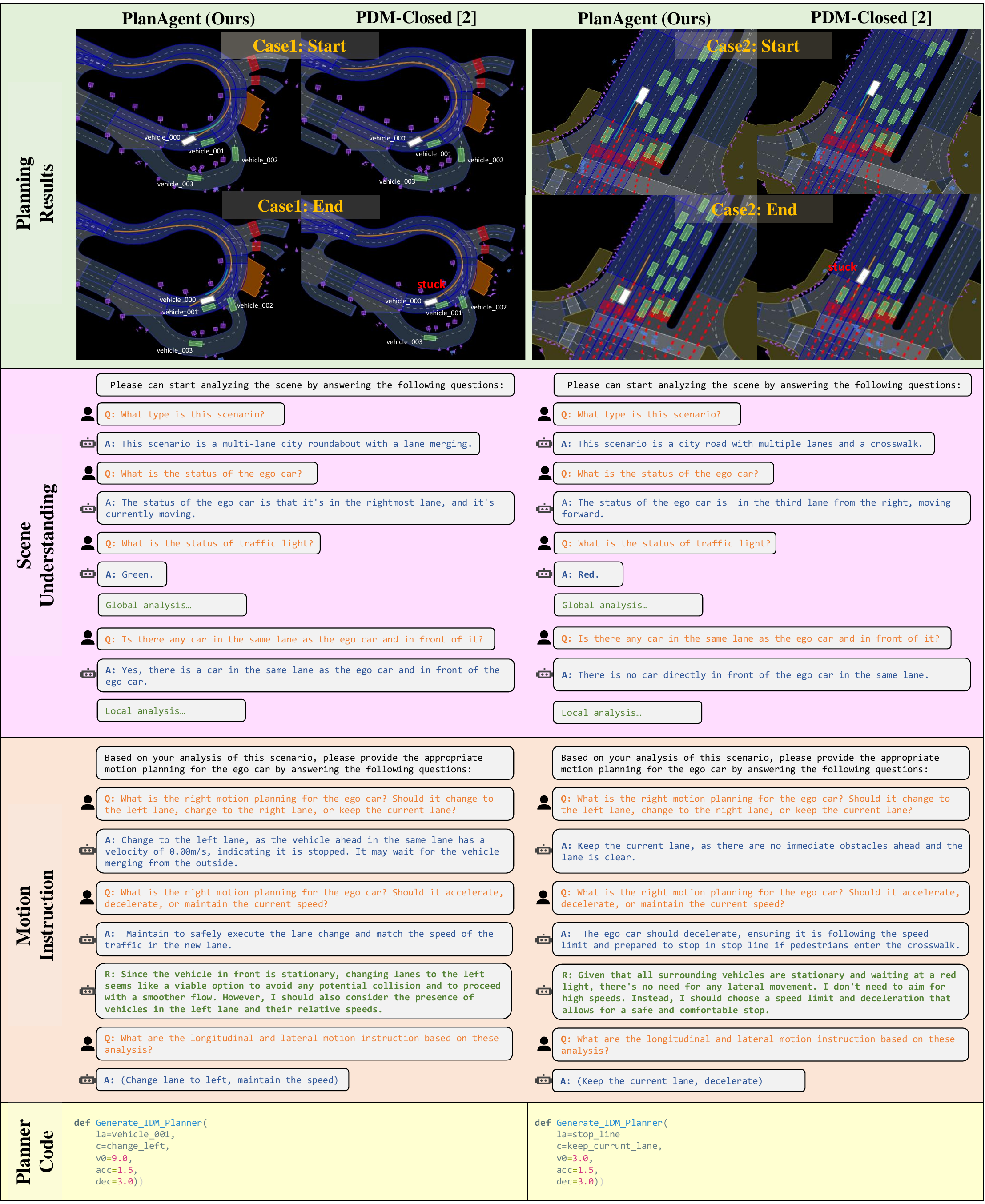}
    \caption{Qualitative comparison between PlanAgent (ours) and PDM-Closed and qualitative example of the hierarchical reasoning of PlanAgent.}
    \label{fig: result}
\end{figure*}

\subsection{Qualitative Results.}
Fig. \ref{fig: result} shows the qualitative results of our PlanAgent comparison with PDM-Closed and the example of PlanAgent's hierarchical reasoning. 

In the first case, the PlanAgent recognizes that it's dealing with a roundabout scenario where vehicles are merging. 
Based on the local motion information of other agents, it is deduced that the vehicle ahead is waiting for vehicles to merge. 
Consequently, PlanAgent provides a lateral instruction to change to the left lane and then generates the corresponding planner code. 
This left lane change allows PlanAgent to avoid being stuck in the middle of the road like PDM-Closed.

In the second case, the PlanAgent recognizes that it's dealing with a scenario with multiple lanes and a crosswalk and the traffic light is red. 
Based on the local motion information of other agents, it infers that the other vehicles are waiting for the red light. 
Thus, PlanAgent provides longitudinal instruction to decelerate for a safe and comfortable stop in the stop line if pedestrians enter the crosswalk. 
This allows PlanAgent to avoid stopping in the middle of the road like PDM-Closed.
\subsection{Interval of MLLM Calls.}
PlanAgent calls the MLLM to understand the scene and generate planner code with a fixed planning interval. 
In Tab. \ref{tab: interval}, we conducted an ablation experiment on the planning interval (frequency) of PlanAgent's MLLM calls for planner generation. 
It is important to note that since the nuPlan simulation frequency is 10 Hz, every 10 planning iterations represent 1 second in the real world. 
The results in Tab. \ref{tab: interval} demonstrate that PlanAgent achieves the best performance at a moderate frequency, specifically with a time interval of 2 seconds. 
Higher or lower call frequencies will hurt the planning performance.

\begin{table}
    \centering
    \setlength{\tabcolsep}{0.04\linewidth}
    \caption{Ablation study of the Interval of MLLM Calls on Test14-hard benchmark.}
    \begin{tabular}[b]{c|cc}
        \toprule
        MLLM Interval (iteration) & 
        NR-CLS $ \uparrow$ & R-CLS $\uparrow$\\
        \midrule
        10 (1s) & 71.85 & 76.10 \\
        20 (2s) & \textbf{72.51} & \textbf{76.82} \\
        30 (3s) & 71.35 & 75.26  \\
        \bottomrule
    \end{tabular}
    \label{tab: interval}
\end{table}

\begin{table}
    \centering
    \setlength{\tabcolsep}{0.027\linewidth}
    \caption{Performance of different Multi-modal Large Language Models for PlanAgent on Test14-hard benchmark.}
    \begin{tabular}[b]{c|cccc}
        \toprule
        MLLM & Param. & Time &
        NR-CLS $\uparrow$ & R-CLS $\uparrow$  \\
        \midrule
         CogVLM-chat\cite{wang2023cogvlm}& 17B & 4405 ms & 68.37 & 72.21 \\
         LLaVa-7B\cite{liu2024visual} & 7B & 2936 ms & 67.73 & 72.06 \\
         LLaVa-13B\cite{liu2024visual} & 13B & 3694 ms & 69.40 & 73.89 \\
         GPT-4V\cite{achiam2023gpt} & - & 5568 ms & 72.51 & 76.82 \\
        \bottomrule
    \end{tabular}
    \label{tab: llm}
\end{table}

\subsection{Performance of Different MLLMs.}
\label{exp_MLLM}

To evaluate the efficacy of various multi-modal large language models in autonomous driving planning tasks, we incorporated widely-used open-source models such as CogVLM-chat\cite{wang2023cogvlm}, LLaVa-7B\cite{liu2024visual}, and LLaVa-13B\cite{liu2024visual} into the PlanAgent system. These models served as MLLM agents within the system.
As shown in Tab. \ref{tab: llm}, LLaVa-13B outperforms the other open-source MLLMs. 
All these three open-source models show significant enhancement over the rule-based method PDM-Closed on the NR-CLS metric, evaluated on the nuPlan Test14-hard benchmark. 
The respective improvements in their driving scores are $5.1\%$, $4.1\%$, and $6.7\%$, illustrating the compatibility of PlanAgent system when working with various MLLM models.
In addition, Tab. \ref{tab: llm} also demonstrates the average time for each MLLM to infer once in the closed-loop simulation.

%% file: a6_limitation.tex
\section{Limitations and Future Work}
\label{conclusion}
\subsection{Limitations.} 
Despite the numerous benefits offered by PlanAgent, it is also subject to the following limitations. Firstly, PlanAgent needs to accurately understand the scenario represented by the BEV map and text description, and generate the planner code based on the understanding of the planner parameters.
Thus, PlanAgent is sensitive to the quality of the prompts that describe scenario and planner parameters. 
However, the current MLLMs such as GPT-4V
and LLaVA are sometimes not accurate enough in scene understanding and planner parameter meaning, leading to the generation of unreasonable planner code. 
Secondly, the current implementation of PlanAgent involves invoking the MLLM to generate the planner code at regular time intervals. 
However, there are instances where the traffic situation in the scene is relatively straightforward, and it does not necessitate the involvement of the MLLM. 
In such cases, rule-based methods such as PDM-Closed\cite{dauner2023parting} can create a safe plan.
Such a fixed-time interval calling MLLM increases the computational burden of closed-loop control.
Lastly, as seen in Tab .\ref{tab: llm}, it takes about 5 seconds to call GPT-4V which does introduce difficulties in deployment. 
We believe it can be solved with the development of the MLLMs.
\subsection{Future Work} 
Considering the aforementioned limitations, we have identified several areas to focus on in our future work. 
Firstly, we will delve into enhancing the MLLM's ability to understand scenarios by fine-tuning it using data collected from autonomous driving datasets. 
This approach aims to improve the accuracy and precision of the MLLM's scene-understanding capabilities.
Secondly, we explore how to determine when MLLM involvement is required, thereby reducing the frequency of MLLM calls and significantly reducing the time and computational resources required for closed-loop planning.
Lastly, we will investigate the utilization of closed-loop simulators to achieve human-like alignment of the MLLM. 
By incorporating feedback from the simulator, we aim to enhance the performance and reliability of the MLLM in real-world driving scenarios.

%% file: a7_conclusion.tex
\section{Conclusion}
\label{conclusion}
In this paper, we present PlanAgent, a novel MLLM-based human-like planning agent system for mid-to-mid vehicle motion planning. 
PlanAgent introduces an Environment Transformation module to  extract BEV map and efficiently generate lane-graph-based textual description as input, proposes a Reasoning Engine module with a well-designed hierarchical chain-of-thought to instruct MLLM to understand driving scenario, give motion instruction, and generate planner code step by step. 
Furthermore, PlanAgent emulates the principles of human decision-making and incorporates a Reflection module, which includes simulation deduction and evaluation scoring, to enhance the safety of planning.
Owing to these enhancements, PlanAgent delivers competitive as well as generalizable results on the nuPlan Val14 and Test14-hard benchmarks and improves the token usage efficiency when describing the driving scenarios. 